\documentclass[journal]{IEEEtran}
\usepackage{graphicx,cite,soul,framed,epsfig,mathptmx,times,amsmath,amssymb,color,flushend,gensymb,subfigure,fancyhdr,subfigure}
\usepackage{lipsum,epstopdf}
\definecolor{shadecolor}{rgb}{1,0.8,0.3}

\ifCLASSINFOpdf
\else
\fi
\begin{document}
\title{Robust Trajectory Tracking Error Model-Based Predictive Control for Unmanned Ground Vehicles}
\author{Erkan~Kayacan,~\IEEEmembership{Student Member, IEEE,} ~~Herman~Ramon~\IEEEmembership{}~and~Wouter~Saeys~\IEEEmembership{}
\thanks{This work was supported by the IWT-SBO 80032 (LeCoPro) project funded by the Institute for the Promotion of Innovation through Science and Technology in Flanders (IWT-Vlaanderen).}
\thanks{E. Kayacan is with the Delft Center for Systems and Control, Delft University of Technology, 2628 CD Delft, The Netherlands.
e-mail: {\tt\small e.kayacan@tudelft.nl }}
\thanks{H. Ramon and W. Saeys are with the Division of Mechatronics, Biostatistics and Sensors, Department of Biosystems, University of Leuven (KU Leuven), Kasteelpark Arenberg 30, B-3001 Leuven, Belgium.
e-mail: {\tt\small \{herman.ramon, wouter.saeys\}@biw.kuleuven.be }}
}
\markboth{\textbf{PREPRINT VERSION:} IEEE/ASME TRANSACTIONS ON MECHATRONICS, Volume 21, Issue 2, 2016.}
{Shell \MakeLowercase{\textit{et al.}}: Bare Demo of IEEEtran.cls for Journals}
\maketitle
\begin{abstract}
This paper proposes a new robust trajectory tracking error-based control approach for unmanned ground vehicles. A trajectory tracking error-based model is used to design a linear model predictive controller and its control action is combined with feedforward and robust control actions. The experimental results show that the proposed control structure is capable to let a tractor-trailer system track both linear and curvilinear target trajectories with low tracking error.
\end{abstract}
\IEEEpeerreviewmaketitle
\section{Introduction}

\begin{figure*}[t!]
\centering
  \includegraphics[width=5.5in]{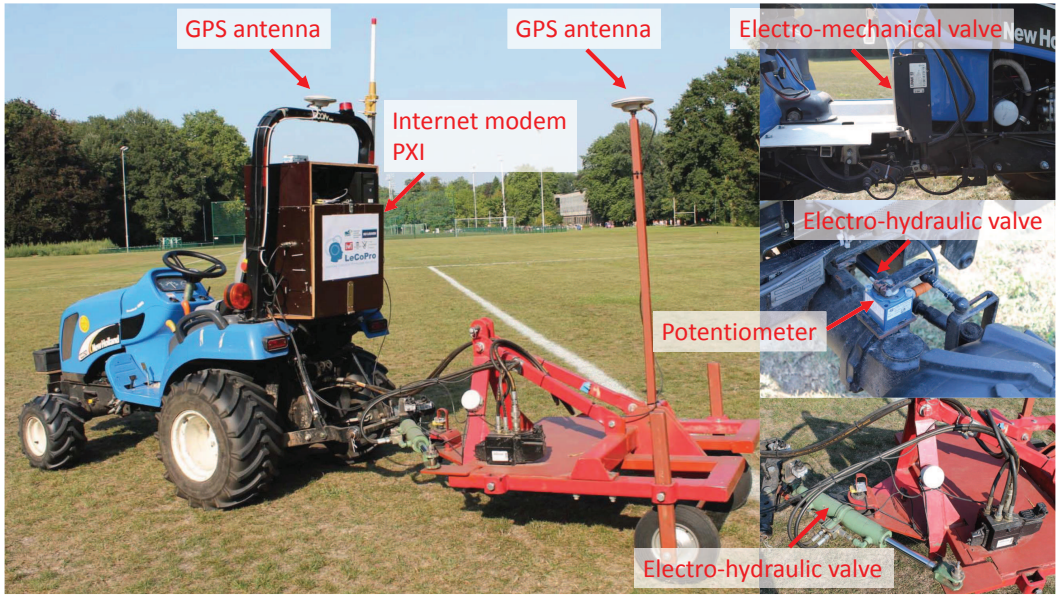}\\
  \caption{The tractor-trailer system}
  \label{tractor-trailer}
\end{figure*}

\IEEEPARstart{T}{hanks} to recent developments in satellite technologies, global positioning system (GPS)-based guidance systems have become very popular. The research on autonomous ground vehicles (AGVs), \emph{e.g.} self-driving cars, has rapidly grown since the introduction of real-time kinematic (RTK) GPS yielding centimeter precision. As automation of agricultural vehicles is essential to remain cost-effective, while they operate at relatively low speed in the field, research on autonomous agricultural vehicles has been increased dramatically after the first successful results on AGVs \cite{Ming2009}.Nowadays, GPS guidance systems on agricultural machinery have become very popular, as they are known to be more accurate than manual navigation, \emph{e.g.} visually straight and parallel crop rows. As a result, the driver no longer has to steer the tractor accurately which is a tiresome task. Moreover, these systems allow to also perform the field work accurately during night or in foggy weather.

The initial studies on autonomous vehicle guidance all used proportional-integral-derivative (PID) controllers. However, the performance of the currently available machine guidance systems controlled by PID is rather limited due to the complex vehicle dynamics which make that the conventional (\emph{e.g.} PID) controllers for machine guidance have to be tuned guardedly or in an adaptive way \cite{Ming2009,wu2001robotica}. Moreover, PID control is a convenient choice for single-input single-output (SISO) systems, while autonomous vehicles have multiple inputs ( \emph{e.g.} speed setting, steering angle setting for the tractor, steering angle setting for the trailer) and outputs (\emph{e.g.}  XY-coordinates of the tractor, XY-coordinates of the trailer, longitudinal speed, yaw rates, yaw angles). These multi-input multi-output (MIMO) systems are traditionally controlled in a decentralized way by designing a controller for each SISO subsystem, thus neglecting the interactions. As an alternative method to PID controllers, optimal control approaches, such as the linear quadratic regulator have been proposed, which are convenient control methods for MIMO systems \cite{Bevly2002,karkeejournal}. Furthermore, model predictive control (MPC) has been suggested as an evolution of the optimal control approach to deal with constraints on the states and the inputs.

In mobile robot applications, successful results for linear MPC (LMPC) were reported when mobile robots are close to the reference. Controllers are generally designed based-on the derived trajectory tracking error-based model, which is a linearized error dynamics model obtained around the reference trajectory, and the control inputs are generally obtained by the combination of feedback and feedforward actions as in \cite{Lee2001,Dongbing2006,Barreto2014}. The LMPC generates the feedback action and the feedforward action is calculated from the reference trajectory \cite{Klancar2007}. On the other hand, in vehicle guidance applications, it has been reported that LMPC worked well for straight line tracking, while no linear control method worked well for curvilinear trajectories \cite{snider2009}. Moreover, these methods cannot achieve trajectory tracking when the system stays off-track and also vehicles are not capable of staying on-track when a curvilinear line starts. The reason is that LMPC works fine for processes which stay around fixed operating-points, as this allows linearization of the process model. However, since the autonomous vehicle has time-varying set points and is subject to several disturbances (e.g., varying soil conditions, bumpy fields), local linearization is not feasible. Moreover, since the model mismatch increases when the system is getting far away from the reference trajectory, it can generate large prediction errors with a consequent instability of the closed-loop system \cite{Falcone2007}. Therefore, nonlinear MPC (NMPC) was proposed as a better alternative \cite{snider2009}.

All aforementioned studies are related to vehicle guidance only. Numerous studies have been reported about the control of vehicle with towed trailer systems such as a tractor-trailer system. The mathematical model of a tractor-trailer system was studied in \cite{karkeejournal}. Three different models were obtained and a linear quadratic regulator based on the linearized model was designed to control the system. It was reported that if the heading angle is more than $10$ degrees, the linearized model is not valid anymore. Moreover, no control law was proposed for the control of the position of the trailer. In \cite{werner2013}, another linear quadratic regulator was proposed for both the control of the tractor and trailer, and thus the position of the trailer was controlled actively. The controller gave successful results for straight line trajectories. However, it was noted that a feedforward control action was required for curved lines tracking. An NMPC implementation for a tractor with a steerable trailer was studied in \cite{Backman2012}. It was reported that the NMPC was able to control the tractor and trailer for straight and also curvilinear lines. Moreover, the system model was made adaptive to varying soil conditions by adding slip parameters for the tractor-trailer system to take the variability in the working environment into account. The nonlinear moving horizon estimator and nonlinear model predictive controller were designed based on the adaptive model in a centralized case and successful experimental results have been reported \cite{erkanCeNMPC}. In addition to centralized NMPC, decentralized and distributed NMPC approaches have respectively been proposed in \cite{erkanDiNMPC,erkanDeNMPC} to decrease the computation time. The experimental results show that although the trajectory tracking accuracy was a little bit worse than the one for centralized NMPC, these approaches reduce the computational cost significantly. Passive control of vehicles with multiple trailers was studied in \cite{Michalek2014}.

Although tracking performance obtained by NMPC was quite good, the computational burden of NMPC implementations is expensive. On the other hand, LMPC is not capable to track curvilinear trajectories accurately although the required computation time is low. The main motivation of this study is to design a robust trajectory-tracking error-based linear model predictive controller for tracking straight and curved lines, and to benchmark its performance in terms of tracking error and computation time against the aforementioned NMPC studies.

This paper is organized as follows: The real-time system and the system model are presented in Section \ref{sectionsystem}. The trajectory tracking error-based model is derived in Section \ref{sectionttebmodel}. In Section \ref{sectionthetrajectorytrackingcontroller}, the feedback control action as an MPC, the feedforward and robust control actions are designed, and the control scheme is presented. The experimental results are presented in Section \ref{sectionexperimentalresults}. Finally, the main conclusions from this study are presented in Section \ref{sectionconc}.



\section{Autonomous Tractor-trailer System and Kinematic Tricycle Model}\label{sectionsystem}
The objective in this study is to track a time-based trajectory with the small agricultural tractor-trailer system shown in Fig. \ref{tractor-trailer}. In practice, an accurate trajectory tracking is desired to  obtain a constant distance between rows to avoid crop damage while difficult and varying soil conditions are faced by a bumpy and wet grass field. The experimental set-up is the same as in \cite{erkanCeNMPC,erkanDeNMPC,erkanDiNMPC}, but the target trajectory is a time-based one instead of a space-based one.

RTK GPS (AsteRx2eH, Septentrio Satellite Navigation NV, Belgium) is used to obtain positional information. For this purpose, two GPS antennas are located straight up the center of the tractor rear axle and the center of the trailer. A Digi Connect WAN 3G modem is used to send uncorrected and receive corrected GPS data from the Flepos network. The non-Gaussian measurement errors of the GPS are $0.03$ m according to the specifications of the manufacturer.

The steering mechanisms of the tractor and trailer consist of electro-hydraulic valve actuators (OSPC50-LS/EH-20, Dan- foss, Nordborg, Denmark) and the speed of the tractor-trailer system is controlled through an electromechanical actuator (LA12, Linak, Nordborg, Denmark) connected to the hydrostat pedal (HP) as shown in Fig. 1. The angle of the front wheels of the tractor is measured using a potentiometer (533-540- J00A3X0-0, Mobil Elektronik, Langenbeutingen, Germany) mounted on the front axle while the steering angle of the trailer is measured by an inductive sensor. The measurements of the steering angles were found to be perturbed by Gaussian noise with standard deviations of 1 degree. An encoder mounted on the rear wheels is used to measure the speed of the system with a measurement error (standard deviation) of $0.1$ $m/s$.

The GPS receiver, the internet modem, all actuators and sensors are connected to a real time operating system (PXI-8110, National Instruments, Austin, TX, USA) through an RS232 serial communication. The PXI system equipped with a 2.26 GHz Intel Core 2 Quad Q9100 quad-core processor acquires all measurements, and controls the tractor-trailer system by applying voltages to the actuators. A laptop is connected to the PXI system by WiFi functions as the user interface of the autonomous tractor-trailer. The control algorithms are imple- mented in $LabVIEW^{TM}$ (version 2011, National Instrument, USA). They are executed in real time on the PXI and updated at a rate of 5-Hz.

\begin{figure}[t!]
\centering
  \includegraphics[width=2.5in]{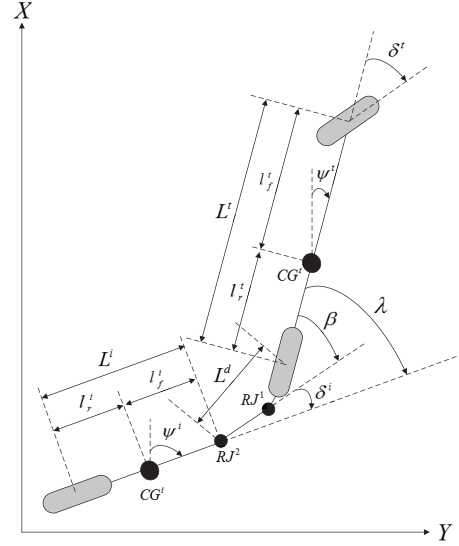}\\
  \caption{Schematic illustration of tricycle model for an autonomous tractor-trailer system}
  \label{kinematic}
\end{figure}

The autonomous tractor-trailer system model is a \emph{kinematic model} neglecting the dynamic force balances in the equations of motion \cite{karkeejournal}. The yaw angle difference between the tractor and the trailer $\lambda$ is defined as the measured relative angle. The tractor and trailer rigid bodies are mechanically linked to each other by the drawbar. There are two revolute joints (RJs) which connect the drawbar to the tractor at $RJ^1$ and the drawbar to the trailer at $RJ^2$ as illustrated in Fig. \ref{kinematic}. The centers of gravity of the tractor and trailer are respectively represented by $CG^t$ and $CG^i$.

The equations of motion of the system consisting of the kinematic and speed models as derived respectively in \cite{erkanCeNMPC,erkanmodelleme} are written as follows:
\begin{eqnarray}\label{kinematicmodel}
  \dot{x}^{t} & = &   v \cos{(\psi^{t})} \nonumber \\
  \dot{y}^{t} & = &  v \sin{(\psi^{t})} \nonumber \\
  \dot{\psi}^{t} & = &\frac{ v \tan{ ( \delta^{t}) } }{L^t} \nonumber \\
  \dot{x}^{i}  & = &  v \cos{(\psi^{i})} \nonumber \\
  \dot{y}^{i} & = &  v \sin{(\psi^{i})} \nonumber \\
  \dot{\psi}^{i}  & = &\frac{ v }{L^i}\big(\sin{( \lambda)} + \frac{l }{L^t} \tan{ ( \delta^{t})} \cos{(\lambda)} \big) \nonumber \\
  \dot{v} & = &-\frac{v}{\tau} + \frac{K}{\tau} HP
\end{eqnarray}
where ${x}^{t}$ and $y^{t}$ represent the position of the tractor, $\psi^{t}$ is the yaw angle of the tractor, ${x}^{i}$ and $y^{i}$ represent the position of the trailer, $\psi^{i}$ is the yaw angle of the trailer, $v$ is the longitudinal speed of the system. Since the tractor and trailer rigid bodies are linked by two RJs at a hitch point, the tractor and the trailer longitudinal velocities are coupled to each other.  The steering angle of the front wheel of the tractor is represented by $\delta^{t}$, $\beta$ is the angle between the tractor and the drawbar at $RJ^{1}$, $\delta^{i}$ is the steering angle between the trailer and the drawbar at $RJ^{2}$, and $HP$ is the hydrostat position. The angle between the tractor and trailer $\lambda$ is equal to the summation of the angle between the tractor and the drawbar at $RJ^{1}$, and the steering angle between the trailer and the drawbar at $RJ^{2}$, ($\lambda= \beta + \delta^{i}$).

The physical parameters that can be directly measured are as follows: The distance between the front axle of the tractor and the rear axle of the tractor $L^t (1.4 m)$, the distance between $RJ^2$ and the rear axle of the trailer $L^i (1.3 m)$, and the distance between the rear axle of the tractor and $RJ^2$ $L^{d} (1.1 m)$, respectively. For an engine speed of $2500$ RPM, the identified parameters are as follows \cite{erkanmodelleme}: the time-constant $\tau=2.05$ and the gain value $K=1.4$ for the speed model.

In the rest of the paper, we denote equations \eqref{kinematicmodel} as
\begin{equation}\label{f_function}
\dot{z}  =  f \big(z,u \big) \\
\end{equation}
where the state, input and output vectors are denoted as follows:

\begin{eqnarray}\label{xstate}
z & = & \left[
  \begin{array}{ccccccc}
   x^{t} & y^{t} & \psi^{t} & x^{i} & y^{i} & \psi^{i} & v
  \end{array}
  \right]^{T} \\ \label{uinput}
u & = & \left[
  \begin{array}{ccc}
   \delta^t & \lambda & HP
  \end{array}
  \right]^T \\ \label{youtput}
y & = & \left[
  \begin{array}{ccccccccc}
   x^{t} & y^{t} & x^{i} & y^{i} & v
  \end{array}
  \right]^{T}
\end{eqnarray}

\section{Trajectory Tracking Error-based System Model}\label{sectionttebmodel}

The trajectory tracking problem is a nonlinear control problem in nature. Therefore, the trajectory tracking control of an autonomous ground vehicle, e.g. tractor-trailer system, can be asymptotically stabilized by nonlinear feedback controllers. In case of linearization around the trajectory, a linear time-varying trajectory tracking system is obtained, which can be controlled by linear controllers \cite{dewit1992,samson1993}. In this section, a new trajectory tracking error-based model is derived. The traditional trajectory tracking error-based models were derived for mobile robots in \cite{Klancar2007} and for the trajectory planner of a tractor-trailer mobile robot in \cite{Khalaji2014}. The difference between the traditional trajectory tracking error-based model and the method proposed here is that the speed and yaw rates are the inputs for the traditional one, while the speed and yaw models are taken into account to design a controller for the new one. As a result, the gas pedal position and steering angles are the inputs for the new trajectory tracking error-based model.

The reference frame represents the inertial reference frame fixed to the motion ground. The other reference frames are moving frames attached to the centers of gravity of the tractor and trailer, which can only translate with respect to the reference frame fixed to the motion ground. The reference trajectory is described by a reference state vector $z_{r} = (x^{t}_{r}, y^{t}_{r}, \psi^{t}_{r}, x^{i}_{r}, y^{i}_{r}, \psi^{i}_{r},  v_{r})^{T}$ and a reference control vector $u_{r} = (\delta^t_{r}, \lambda_{r}, HP_{r})^{T}$. The error state $z_{e}$ expressed in the frames on the tractor and trailer is written as follows:
\begin{eqnarray}\label{errorstate}
z_{e}  =  T \times [z_{r} - z]
\end{eqnarray}
where $T$ is the transformation matrix between reference frames as follows:
\begin{equation}
 \left[
\begin{array}{ccccccc}
\cos{(\psi^{t})} & \sin{(\psi^{t})} & 0 & 0 & 0 & 0 & 0  \\
-\sin{(\psi^{t})} & \cos{(\psi^{t})} & 0 & 0 & 0 & 0 & 0  \\
0 & 0 & 1 & 0 & 0 & 0 & 0 \\
0 & 0 & 0 & \cos{(\psi^{i})} & \sin{(\psi^{i})} & 0 & 0  \\
0 & 0 & 0 & -\sin{(\psi^{i})} & \cos{(\psi^{i})} & 0 & 0  \\
0 & 0 & 0 & 0 & 0 & 1 & 0 \\
0 & 0 & 0 & 0 & 0 & 0 & 1 \\
\end{array}
 \right] \nonumber
\end{equation}

The trajectory tracking error-based model is derived by taking the derivative of the error state in \eqref{errorstate} and taking the system model in \eqref{kinematicmodel} into account as follows:
\begin{eqnarray}\label{errormodel}
\dot{x}^{t}_{e}  & = & \gamma^{t} y^{t}_{e} - v + v_{r} \cos{(\psi^{t}_{e})} \nonumber \\
\dot{y}^{t}_{e}  & = & - \gamma^{t} x^{t}_{e} + v_{r} \sin{(\psi^{t}_{e})} \nonumber \\
\dot{\psi}^{t}_{e} & = & \frac{v_r \tan{(\delta^t_{r})} - v \tan{(\delta^t)} }{L^{t}} \nonumber \\
\dot{x}^{i}_{e}  & = & \gamma^{i} y^{t}_{e} - v + v_{r} \cos{(\psi^{i}_{e})} \nonumber \\
\dot{y}^{i}_{e}  & = & - \gamma^{i} x^{t}_{e} + v_{r} \sin{(\psi^{i}_{e})} \nonumber \\
\dot{\psi}^{i}_{e}   & = & \frac{v_{r} }{L^i}\big(\sin{(\lambda_{r})} + \frac{L^d }{L^t} \tan{( \delta^{t}_{r})} \cos{(\lambda_{r})} \big) \nonumber \\
                  && -\frac{v }{L^i}\big(\sin{(\lambda)} + \frac{L^d }{L^t} \tan{( \delta^{t})} \cos{(\lambda)} \big) \nonumber \\
\dot{v}_{e}      & = & - \frac{v_{e}}{\tau} + \frac{K}{\tau} ( HP_{r} - HP )
\end{eqnarray}
where $\gamma^{t}$ and $\gamma^{i}$ are the yaw rates of the tractor and trailer, respectively.

The trajectory tracking error-based model in the state-space form is written by linearizing the error model in \eqref{errormodel} around the reference trajectory ($x^{t}_{e}=y^{t}_{e}=\psi^{t}_{e}=x^{i}_{e}=y^{i}_{e}=\psi^{i}_{e}=v_{e}=\delta^{t}_{e}=\lambda_{e}=HP_{e}=0$) as follows:
\begin{eqnarray}\label{errorstateupdate}
\dot{z}_{e} & = & A z_{e} + B u_{e} \nonumber \\
\dot{z}_{e} & = & \left[
                \begin{array}{ccccccc}
                  0 & \gamma^{t}_{r} & 0 & 0 & 0 & 0 & 1  \\
                  -\gamma^{t}_{r} & 0 & v_{r} & 0 & 0 & 0 & 0  \\
                  0 & 0 & 0 & 0 & 0 & 0 & 0 \\
                  0 & 0 & 0 & 0 & \gamma^{i}_{r} & 0 & 1  \\
                  0 & 0 & 0 & -\gamma^{i}_{r} & 0 & v_{r} & 0  \\
                  0 & 0 & 0 & 0 & 0 & 0 & 0 \\
                  0 & 0 & 0 & 0 & 0 & 0 & -\frac{1}{\tau} \\
                \end{array}
              \right] z_{e} \nonumber \\
              && +
              \left[ \begin{array}{ccc}
                  0 & 0 & 0  \\
                  0 & 0  & 0 \\
                  \frac{v_{r}}{L^{t}} & 0 & 0   \\
                  0 & 0 & 0   \\
                  0 & 0 & 0   \\
                  \frac{v_{r} L^{d}}{L^{t} L^{i}} & \frac{v_{r}}{L^{i}} & 0   \\
                  0 & 0 & \frac{K}{\tau}   \\
                \end{array}\right] u_{e}
\end{eqnarray}
where the state and control vectors for the trajectory tracking error-based model are denoted as
\begin{eqnarray}\label{xstate}
z_{e} & = & \left[
  \begin{array}{ccccccc}
   x^{t}_{e} & y^{t}_{e} & \psi^{t}_{e} & x^{i}_{e} & y^{i}_{e} & \psi^{i}_{e} & v_{e}
  \end{array}
  \right]^{T} \\ \label{uinput}
u_{e} & = & \left[
  \begin{array}{ccc}
   \delta^t_{e} & \lambda_{e} & HP_{e}
  \end{array}
  \right]^T
\end{eqnarray}

Remark 1: The trajectory tracking error-based model is controllable when either the reference longitudinal velocity $v_{r}$ or the reference yaw rates $\gamma^{t}_{r}$ and $\gamma^{i}_{r}$ are nonzero, which is a sufficient condition.

\section{Design of The Robust Trajectory Tracking Error-Based Controller} \label{sectionthetrajectorytrackingcontroller}

\begin{figure*}[t!]
\centering
  \includegraphics[width=7in]{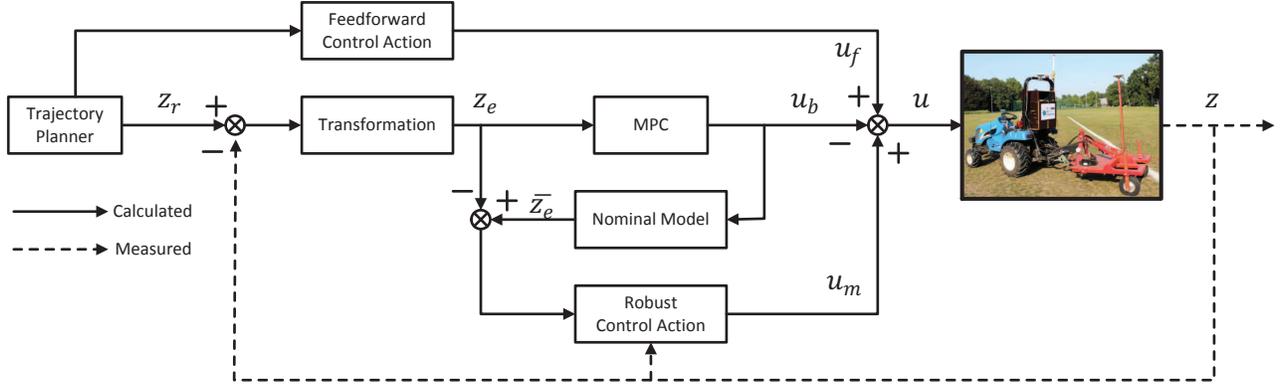}\\
  \caption{Block diagram of the control scheme combining feedback MPC, feedforward and robust control actions.}
  \label{controlscheme}
\end{figure*}

The control scheme is illustrated in Fig. \ref{controlscheme}. The control input applied to the real-time system is calculated as the difference between the summation of the feedforward $u_{f}$ and robust $u_{m}$ control actions, and the feedback $u_{b}$ control action:

\begin{equation}
u = u_{f} - u_{b} + u_{m}
\end{equation}

In following subsections, the feedback, feedforward and robust control actions are formulated.

\subsection{Feedback Control Action: Model Predictive Control}\label{sectionmpc}

The system to be controlled is described by the following linear discrete-time model:
\begin{equation}
z_{e}(k+1) = A_{d} z_{e}(k) + B_{d} u_{e}(k)
\end{equation}
where $z_{e} (k)$ $\in$ $\mathbb{R}^{n_{z}}$ is the state vector and $u_{e} (k)$ $\in$ $\mathbb{R}^{n_{u}}$ is the control input. The matrices $A_{d}$ and $B_{d}$ are calculated considering the sampling time of the real-time system by using the continuous-time version of the trajectory tracking error-based model in \eqref{errorstateupdate}.

The constraints are written for all $k \geq 0$ as follows:
\begin{eqnarray}\label{constraints}
-55 \; degrees/s  \leq & \Delta \delta^{t}_{e}(k) & \leq  55 \; degrees/s  \nonumber \\
-35 \; degrees/s  \leq & \Delta \lambda_{e}(k) & \leq  35 \; degrees/s \nonumber \\
-30 \; \% /s \leq & \Delta HP_{e}(k) & \leq  30 \; \% /s \nonumber \\
-12 \; degrees  \leq &  \delta^{t}_{e}(k) & \leq  12 \; degrees  \nonumber \\
-6 \; degrees  \leq &  \lambda_{e}(k) & \leq  6 \; degrees \nonumber \\
-10 \; \%  \leq &  HP_{e}(k) & \leq  10 \; \%
\end{eqnarray}

The cost function in its general form is written as follows:
\begin{equation}\label{costfunction}
J\big(\Delta U, z_{e}(k)\big)=\displaystyle\sum\limits_{i=0}^{N_{p}} z^{T} _{e_{k+i|k}} Q z_{e_{k+i|k}} + \displaystyle\sum\limits_{i=0}^{N_{c}-1} \Delta u^{T} _{e_{k+i}} R \Delta u_{e_{k+i}}
\end{equation}
where $N_p=8$ and $N_c=3$ represent the prediction and control horizons, $\triangle u_{e}$ is the change of the input, and $\Delta U = [\Delta u^{T} _{e_{k}},...,\Delta u^{T} _{e_{k+N_{c}-1}}]^{T}$ is the vector of the input steps from sampling instant $k$ to sampling instant $k+N_{c}-1$. Since the sampling time of the real-time experiments has been equal to $200$ ms, the prediction and control horizons are respectively equal to $1.6$ s and $0.6$ s. The positive-definite weighting matrices $Q^{n_{z} \times n_{z}}$ and $R^{n_{u} \times n_{u}}$ are defined as follows:
\begin{equation}\label{}
Q  =  diag(1,1,0,1,1,0,0) \;\;\; , \;\;\; R  =  diag(1,1,1)
\end{equation}

Since the aim is trajectory tracking control of the tractor-trailer system, we try to minimize the tractor and trailer tracking errors on x- and y-axes. If oscillatory behavior is observed, then the yaw angle error for the lateral motion and the speed error for the longitudinal motion might be needed to minimize. On the other hand, if the values for the yaw angle and speed errors are set to very large values, the system may not be able to track the reference trajectory. Since we have not observed any oscillatory behavior, we have not needed to minimize the yaw angle and speed errors.

The following plant objective function is solved at each sampling time for the LMPC:
\begin{equation}
 \begin{aligned}
 & \underset{z_{e}(.), u_{e}(.)}{\text{min}}
 & & \displaystyle\sum\limits_{i=0}^{N_{p}} z^{T} _e{_{k+i|k}} Q z_{e_{k+i|k}} + \displaystyle\sum\limits_{i=0}^{N_{c}-1} \Delta u^{T} _{e_{k+i}} R \Delta u_{e_{k+i}}  \\
 & \text{subject to}
 && z_{e}(k+1) = A_{d} z_{e}(k) + B_{d} u_{e}(k) \\
 &&&  -55 \; degrees/s  \leq \Delta \delta^{t}_{e}(k) \leq  55 \; degrees/s  \\
 &&&  -35 \; degrees/s  \leq \Delta \lambda_{e}(k) \leq  35 \; degrees/s  \\
 &&&  -30 \leq \Delta HP_{e}(k) \leq  30  \\
 &&& -12 \; degrees  \leq   \delta^{t}_{e}(k)  \leq  12 \; degrees   \\
 &&& -6 \; degrees  \leq   \lambda_{e}(k)  \leq  6 \; degrees  \\
 &&& -10 \; \%  \leq   HP_{e}(k)  \leq  10 \; \%
  \end{aligned}
  \label{mpc}
\end{equation}

In the real-time implementation of the LMPC, the linear optimization problem in \eqref{mpc} is solved online for a given $z_{e}(k)$ in a receding horizon fashion. In this approach, the first element of the input sequence is applied to the system, while the rest is discarded. For the next time step, the entire procedure is repeated for the new measured or estimated output. The online MPC algorithm can be implemented in the following steps:
\begin{enumerate}
  \item Measure or estimate the current system states $z_{e}(k)$
  \item Solve the optimization problem in \eqref{mpc} to obtain $\triangle U^*=[\triangle u^*_{e} (k),\ldots,\triangle u^*_{e} (k+N_{c}-1)]^{T}$
  \item Apply $u^*_{e} (k) = \triangle u^*_{e} (k) + u^*_{e} (k-1) $
\end{enumerate}
The optimization problem is then solved over a shifted horizon for the next sampling time.

In our case, the designed LMPC minimizes the differences between the reference trajectory and the measured positions of the tractor and trailer in x- and y-axes, and finds the differences between the reference and actual control inputs. For this reason, the generated inputs by the LMPC are not the actual control inputs to the real-time system. Therefore, we will define feedforward control actions in the next subsection \ref{sectionfeedforwardcontrolaction} to calculate the control inputs applied to the real-time system. The input calculated by the LMPC $u^{*}_{e}$ contributes to the input signal to the system $u$ as a feedback control action $u_{b}$:
\begin{equation}\label{ue}
u_{b}= u^*_{e}
\end{equation}

Remark 2: As can be seen from \eqref{mpc}, the equality and inequality constraints are linear such that the formulation is a convex optimization problem. A quadratic programming solver can be used for this optimization problem. The formulation for NMPC  is the constrained nonlinear optimization problem which is non-convex. For this reason, it is to be noted that the computational burden of the optimization problem for LMPC in \eqref{mpc} is significantly lower than the one for NMPC.

Once the control input applied to the real-time system has been found, a modification is required to find the steering angle for the trailer $\delta^{i}$. This is found by subtracting the angle $\beta$ between the tractor and the drawbar at $RJ^{1}$ from the angle $\lambda$ between the tractor and the trailer  as follows:
\begin{equation}
\delta^{i} = \lambda - \beta
\end{equation}

\subsection{Feedforward Control Action}\label{sectionfeedforwardcontrolaction}

As the LMPC generates the differences between the reference and actual control variables, the outputs of the MPC have to be combined with a feedforward control action to calculate the actual control inputs to be applied to the real-time system. Feedforward control inputs $\delta^{t}_{r}$, $\lambda_{r}$ and $HP_{r}$ are derived for given reference trajectories ($x^{t}_{r}$, $y^{t}_{r}$, $x^{i}_{r}$, $y^{i}_{r}$) by using the system model in \eqref{kinematicmodel}. The feedforward control actions for the tractor-trailer system are the reference steering angles and the reference hydrostat position. The reference longitudinal velocity $v_{r}$, and the reference yaw rates $\gamma^{t}_{r}$ and $\gamma^{i}_{r}$  for the tractor-trailer system are derived for a given reference trajectory $(x^{t}_{r}, y^{t}_{r}, x^{i}_{r}, y^{i}_{r})$ defined in a time interval $t \in [0,T]$ as follows:
\begin{eqnarray}
v_{r} & = & \pm \; \sqrt[]{ (\dot{x}^{t}_{r})^{2} + (\dot{y}^{t}_{r})^{2} } \nonumber \\
\gamma^{t}_{r} & = & \frac {\dot{x}^{t}_{r} \ddot{y}^{t}_{r} - \dot{y}^{t}_{r} \ddot{x}^{t}_{r} } {(\dot{x}^{t}_{r})^{2} + (\dot{y}^{t}_{r})^{2}} \nonumber \\
\gamma^{i}_{r} & = & \frac {\dot{x}^{i}_{r} \ddot{y}^{i}_{r} - \dot{y}^{i}_{r} \ddot{x}^{i}_{r} } {(\dot{x}^{i}_{r})^{2} + (\dot{y}^{i}_{r})^{2}}
\end{eqnarray}
where the sign $\pm$ indicates the desired driving direction of the system ($+$ for forward, $-$ for reverse).

To calculate the feedforward control action for the tractor-trailer system, the steering angles are assumed to be small, and the steady-state behaviour of the relation between the longitudinal velocity and the hydrostat position is taken into account. Under these assumptions, the feedforward control actions can be derived from \eqref{kinematicmodel} as:
\begin{eqnarray}\label{ffcontrolaction}
\delta^{t}_{r} & = & \frac{\gamma^{t}_{r} L^{t} }{v_{r}}  \nonumber \\
\delta^{i}_{r} & = & \frac{\gamma^{i}_{r} L^{i} -\gamma^{t}_{r} L^{d} }{v_{r}} - \beta\nonumber \\
HP_{r} & = & \frac{v_{r}}{K}
\end{eqnarray}
The defined feedforward control action $u_{F}= [\delta^{t}_{r}, \delta^{i}_{r}, HP_{r}]^{T}$ provides the calculated references for the control inputs. The calculated feedforward control action will only be able to drive the tractor-trailer system on the reference trajectory if there are no disturbances, uncertainties and initial state errors.

Remark 3: The necessary condition in the trajectory design is that the trajectory is twice-differentiable, and the velocity reference $v_{r}\neq0$ and the gain of the speed model $K\neq0$ are nonzero.

\subsection{Robust Control Action}\label{sectionrobustcontrolaction}

Since the trajectory tracking error-based model has been obtained by linearizing the system around the reference trajectory, the mismatch between the trajectory tracking error-based model and the real system can result in poor control performance when the system is not close to the reference. Therefore, a robust control action is required to bring and to keep the system close to the reference trajectory.

In \cite{Mayne2005,Mayne2006,Maynenonlineartube}, a tube-based approach for (N)MPC was proposed to obtain robust and better control performance of the system. The robust control law is written as follows:
\begin{equation}
u_{m} = K \big(\bar{z}_{e}(t) - z_{e}(t) \big)
\label{robustcontrollaw}
\end{equation}

where $K \in \mathbb{R}^{n_{u} \times n_{z}}$ is the feedback gain and $\bar{z}_{e}(t) - z_{e}(t) $ is the modeling error between the nominal model in \eqref{errorstateupdate} and the real system.

The modeling error term is calculated as the difference between the linearized model and the real system as follows:
\begin{equation}
z_{m} = g \big(z_{e}(t),u(t)\big)
\label{z}
\end{equation}
where $z_{m} \in \mathbb{Z}_{m}$ is a robust positively invariant set. It is assumed that $\mathbb{Z}_{m} \subset \mathbb{Z}_{e}$ and $K \mathbb{Z}_{m} \subset \mathbb{U}$. The nominal state and input have to satisfy:
\begin{eqnarray}
\bar{z}_{e} & \in & \mathbb{\bar{Z}}_{e} =\mathbb{Z}_{e} \ominus \mathbb{Z}_{m} \nonumber \\
\bar{u} & \in & \mathbb{\bar{U}} = \mathbb{U} \ominus K \mathbb{Z}_{m}
 \end{eqnarray}
where they are in the neighborhoods of the origin.

Since only measurable states must be considered in the robust control law due to the estimation error, the uncertainty vector is written as follows:

\begin{eqnarray}
z_{m} = \left[ \begin{array}{c}
            x^{t}_{m} \\
            y^{t}_{m} \\
            x^{i}_{m} \\
            y^{i}_{m}
          \end{array}
\right] = \left[ \begin{array}{c}
             \bar{x}^{t}_{e} - x^{t}_{e} \\
            \bar{y}^{t}_{e} - y^{t}_{e}\\
            \bar{x}^{i}_{e} - x^{i}_{e}\\
            \bar{y}^{i}_{e} - y^{i}_{e}
          \end{array}
\right]
\label{z2}
\end{eqnarray}
where $z_m$ vector is defined as $-1\leq z_m \leq 1$ in this paper.

In most tractor-trailer systems, there are actuator limits. For this reason, the constraints on the actuators must be taken into account. Therefore, we propose a $tanh$ function to place a saturation for the robust control action \cite{erdalt2fnn,erkantowards}. Moreover, during the real-time experiments, it was observed that if only the uncertainty vector was considered, the system exhibited oscillatory behaviour. For this reason, the derivative of the uncertainty vector is also considered to reduce overshoots. This results in the following robust control law for the tractor and trailer:
\begin{eqnarray}\label{robustcontrollaw}
u_{m} =  \left[ \begin{array}{c}
         \delta^{t}_{m} \\
         \delta^{i}_{m} \\
                  HP_{m}
       \end{array} \right]  =  \left[ \begin{array}{c}
         k_{s_{1}} \tanh{(k_{p_1}  y^{t}_{m} + k_{d_1}  \dot{y}^{t}_{m} )} \\
         k_{s_{2}} \tanh{(k_{p_2}  y^{i}_{m} + k_{d_2}  \dot{y}^{i}_{m} )}  \\
         k_{s_{3}} \tanh{(k_{p_3}  x^{t}_{m} + k_{d_3}  \dot{x}^{t}_{m} )}
       \end{array} \right]
\end{eqnarray}
where $k_{p}>0$, $k_{d}>0$ and $k_{s}>0$ are respectively the proportional, derivative and saturation gains of the robust control term. The saturation coefficients have been determined by considering the constraints on the actuators, the feedback controller and the feedforward control action. For this reason, the saturation coefficients $k_{s_{1}}$, $k_{s_{2}}$ and $k_{s_{3}}$ have been respectively set to $0.2$, $0.1$ and $10$. Moreover, the system shows oscillatory behaviour when the derivative coefficients are not larger than the proportional ones. Therefore, the proportional coefficients $k_{p_{1}}$, $k_{p_{2}}$ and $k_{p_{3}}$ have been respectively set to $2$, $1$ and $10$ while the derivative coefficients $k_{d_{1}}$, $k_{d_{2}}$ and $k_{d_{3}}$ have been respectively set to $4$, $2$ and $20$. The robust control actions $\delta^{t}_{m}$, $\delta^{i}_{m}$ and $HP_{m}$ are the steering angles of the tractor and trailer, and the hydrostat position.

The nominal controller $\bar{u}_{e} (t)$ in \eqref{ue} is calculated online, while the ancillary control law $k_{p}$, $k_{d}$ and $k_{s}$ obtained offline keeps the trajectories of the system error in the robust control invariant set $z_{m}$ centered along the nominal trajectory \cite{Mayne2005}. The stability issue of the robust tube-based MPC of constrained linear system with disturbances was clarified in \cite{Mayne2005,Limon2010248}.

\section{Experimental Results}\label{sectionexperimentalresults}

\begin{figure}[h!]
\centering
\includegraphics[width=3.6in]{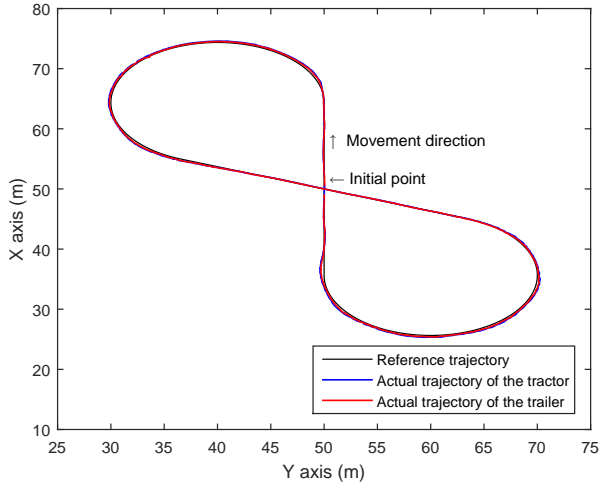}\\
\caption{Reference and actual trajectories}
\label{traj}
\end{figure}

The time-based, 8-shaped trajectory illustrated in Fig. \ref{traj} has been used as the reference signal. The 8-shaped trajectory consists of two straight lines and two smooth curves. Since the radius of the curves is equal to $10$ m, the curvature of the smooth curves is equal to $0.1$. (The curvature of a circle is the inverse of its radius).

\begin{figure}[h!]
\centering
\includegraphics[width=3.6in]{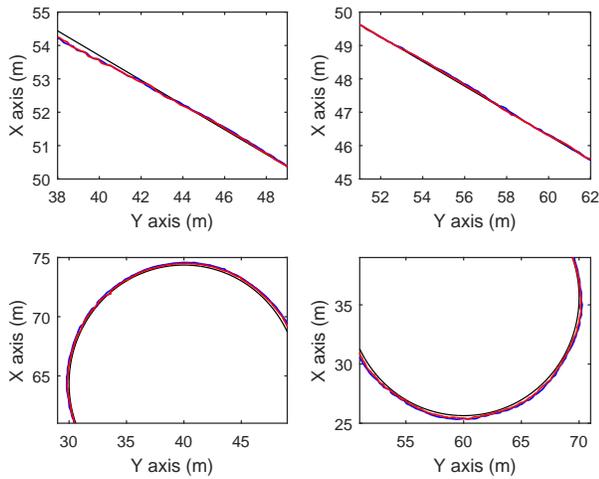}\\
\caption{Zoom versions of trajectories}
\label{zoom_traj}
\end{figure}

The actual trajectories of the tractor and trailer are shown in Fig. \ref{traj}, while close ups are shown in Fig. \ref{zoom_traj}. Thanks to the robust control action, the new robust trajectory tracking-based model predictive controller is capable to navigate the autonomous tractor-trailer system close to the target trajectory. Moreover, the system did not exhibit oscillatory behaviour.
\begin{figure}[h!]
\centering
\includegraphics[width=3.6in]{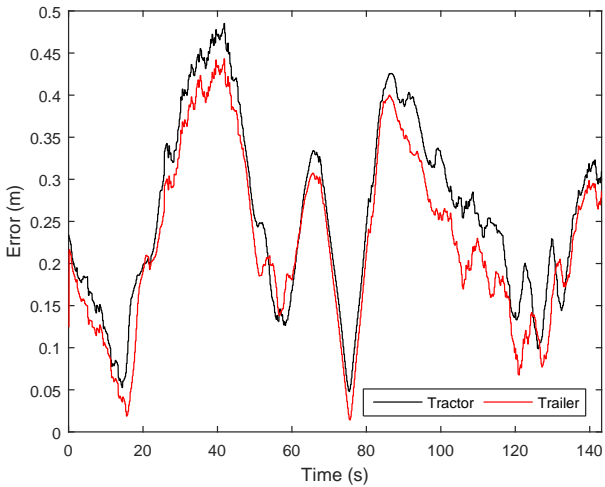}\\
\caption{Euclidian distance error to the reference trajectory}
\label{error}
\end{figure}

The Euclidian distance errors to the time-based reference trajectory for both the tractor and the trailer are shown in Fig. \ref{error}. The mean values of the Euclidian distance errors of the tractor and the trailer for the straight lines are respectively $23.49$ cm and $21.21$ cm. Besides, the mean values of the Euclidian distance errors of the tractor and the trailer for the curved lines are respectively $39.82$ cm and $36.21$ cm. As can be observed, the trajectory tracking error of the tractor-trailer system for straight lines is lower than for the curved lines. NMPC was used for the space-based trajectory approach in \cite{erkanCeNMPC}. It was reported that the Euclidean error values of the tractor and the trailer for the straight lines are respectively  $6.44$ cm and $3.61$ cm, while the Euclidean error values of the tractor and the trailer for the curved lines are respectively $49.78$ cm and $41.52$ cm. As can be observed, the tracking error to the space-based trajectory was less than the one to the time-based trajectory for straight lines, while it was more than the one to the time-based trajectory for curved lines. Therefore, it can be concluded that the preferred approach depends on the shape of the trajectory.

\begin{figure}[t!]
\centering
\includegraphics[width=3.6in]{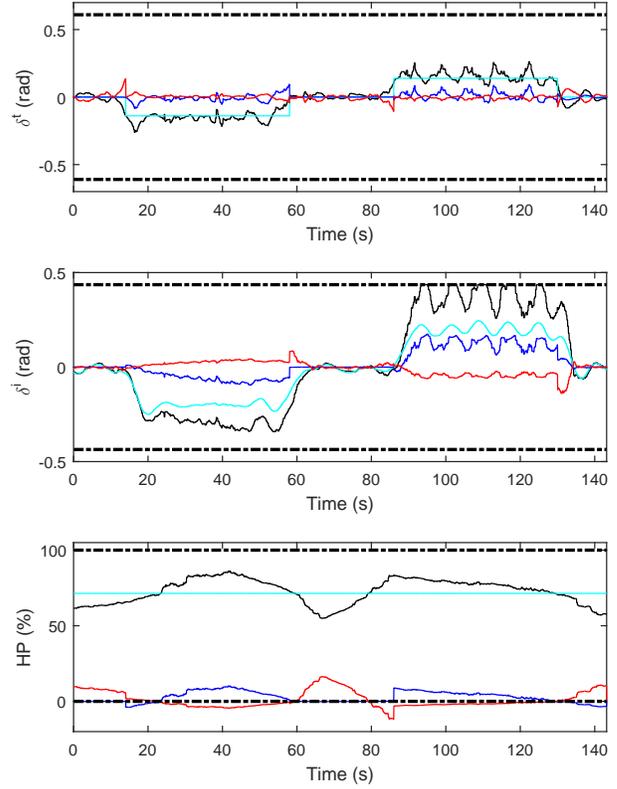}\\
\caption{Control Signals: dashed line: bound for the total control action, black line: total control action, cyan line: feedforward control, red line: feedback control, blue line: robust control.}
\label{controlsignals}
\end{figure}
In Fig. \ref{controlsignals}, the outputs, the steering angle ($\delta^{t}$) reference for the tractor, the steering angle ($\delta^{i}$) reference for the trailer, and the hydrostat position (HP) reference, of the controller  are illustrated. As can be seen from this figure, the total control inputs are within the bounds, and the feedback, feedforward and robust control actions can be observed. The contribution of the robust control action is more than the one of the feedback control action for curved lines, while the one of feedback control action is dominant for straight lines. The reason is that the yaw angle is time-invariant for straight lines, while it is time-varying for curved lines. This results in larger mismatch problem during tracking curved lines. Moreover, the steering angle reference for the trailer has some oscillation. As can be seen in \eqref{ffcontrolaction}, it is calculated considering the reference trajectory and the measured hitch point angle $\beta$. Thus, this oscillatory behaviour is caused by the measured hitch point angle $\beta$.

The average computation time for LMPC was equal to $1.1$ ms and feasible in real-time. As reported in \cite{erkanCeNMPC}, while the computation time for NMPC was still acceptable for real-time applications, the average computation time for NMPC was $6$ times larger with $6.8$ ms. It is to be noted that the computation time increases exponentially when the number of the state and input increases.

\section{Conclusions}\label{sectionconc}

A new robust trajectory tracking error-based model predictive controller has been elaborated for the control of an autonomous tractor-trailer system. To increase the robustness of the algorithm, the tube-based approach has been used, and it was evaluated in real-time with respect to its computation time and tracking accuracy. The experimental results in the field have shown that the designed controller is able to control the system with a reasonable accuracy due to the modeling errors and disturbances. The mean values of the Euclidian distance errors on the straight lines for the tractor and the trailer were respectively equal to $23.49$ cm and $21.21$ cm, while the ones for the curved lines are respectively $39.82$ cm and $36.21$ cm. The computation time for LMPC was around $1.1$ ms and is significantly smaller than for NMPC.

\section*{Acknowledgment}
We would like to thank Mr. Soner Akpinar for his technical support for the preparation of the experimental set up.

\bibliography{ref_ttebmpc}
\bibliographystyle{IEEEtran}

\begin{IEEEbiography}[{\includegraphics[width=1in,height=1.25in,clip,keepaspectratio]{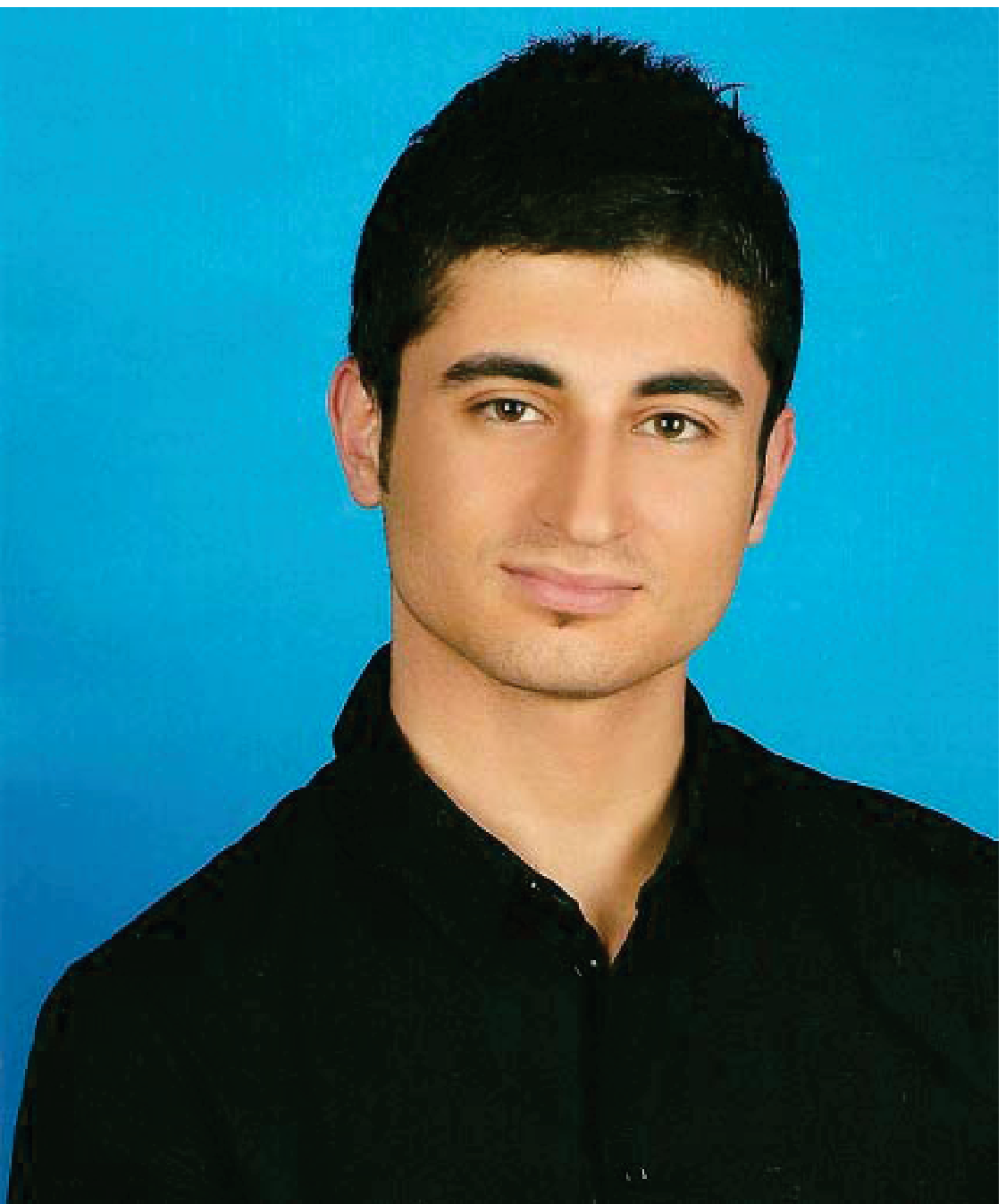}}]{Erkan Kayacan} (S\textquoteright 12) was born in Istanbul, Turkey, on April 17, 1985. He received the B.Sc. and the M.Sc. degrees in mechanical engineering from Istanbul Technical University, Istanbul, Turkey in 2008 and 2010, respectively. He received the Ph.D. degree in Mechatronics, Biostatistics and Sensors from University of Leuven (KU Leuven), Leuven, Belgium in 2014.

He is currently a Postdoctoral Researcher with the Delft Center for Systems and Control, Delft University of Technology, Delft, The Netherlands. His current research interests include model predictive control, state estimation, unmanned vehicles and autonomous systems.
\end{IEEEbiography}

\begin{IEEEbiography}[{\includegraphics[width=1in,height=1.25in,clip,keepaspectratio]{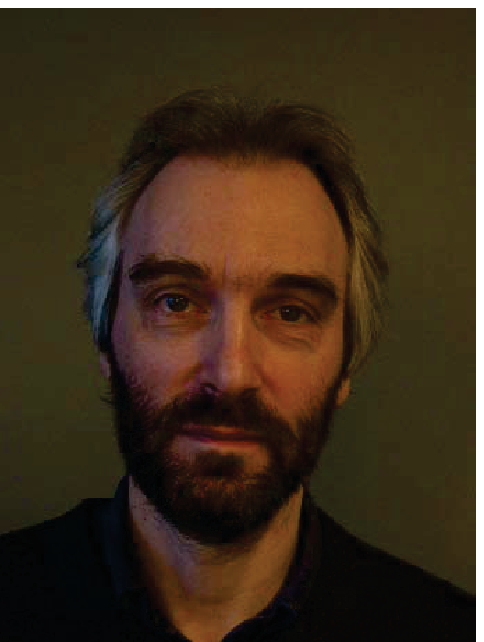}}]{Herman Ramon} received the M.Sc. degree in bioscience engineering from Gent University, Gent, Belgium and the Ph.D. degree in biological sciences from the University of Leuven (KU Leuven), Leuven, Belgium, in 1993.

He is currently a Professor with the Faculty of Bioscience Engineering, KU Leuven, lecturing on field robotics, system dynamics, applied mechanics and mathematical biology. His current research interests include precision technologies and advanced mechatronic systems for processes involved in the production chain of food and non-food materials, from the field to the end user. He has authored and co-authored more than 200 peer reviewed journal articles (ISI).
\end{IEEEbiography}

\begin{IEEEbiography}[{\includegraphics[width=1in,height=1.25in,clip,keepaspectratio]{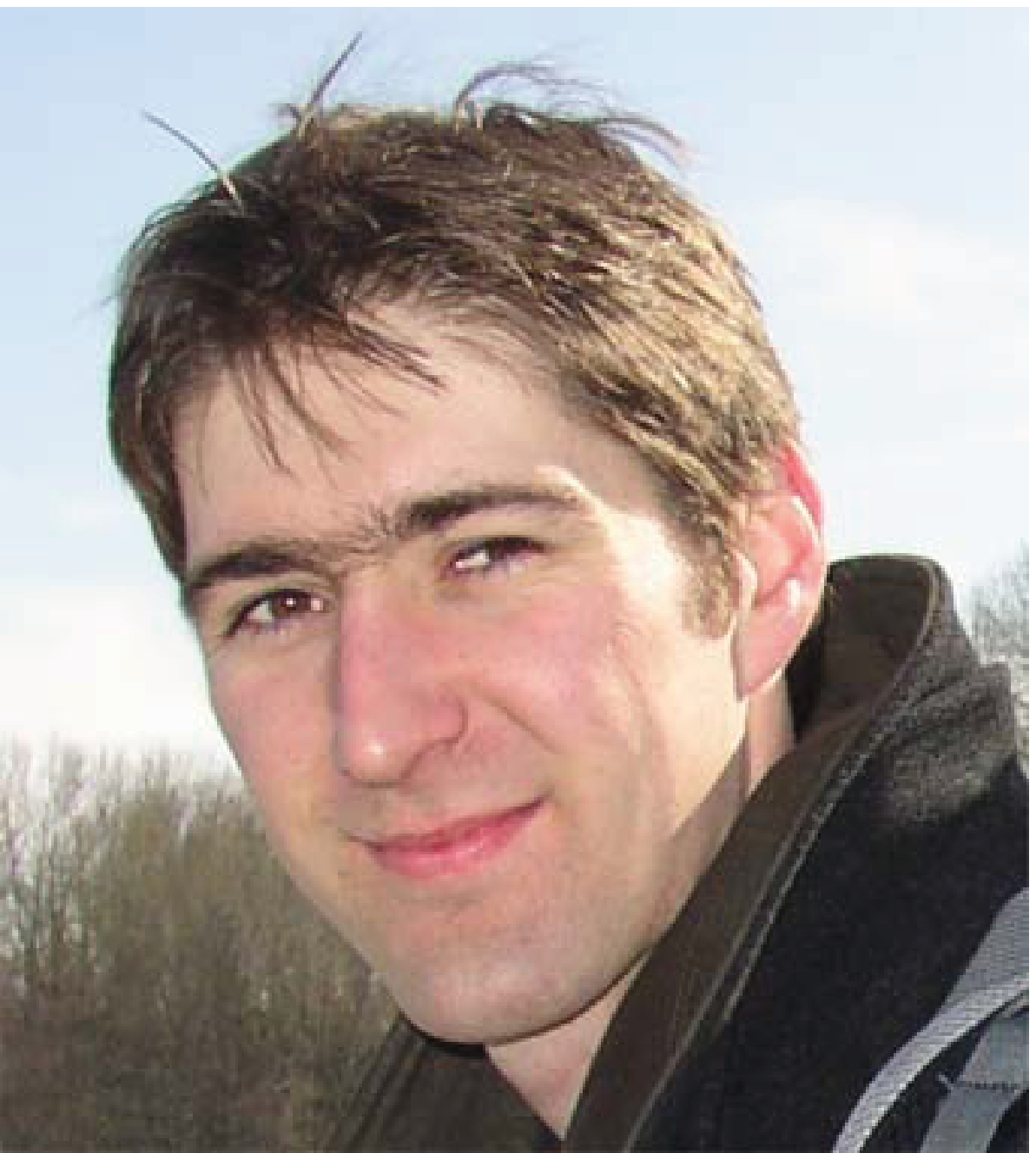}}]{Wouter Saeys} received the M.Sc degree in Bioscience Engineering from University of Leuven (KU Leuven), Leuven, Belgium in 2002. On the basis of his Master’s thesis, he was awarded  the engineering prize by the Royal Flemish Society of Engineers (KVIV). In 2006, he received the Ph.D. in Bioscience Engineering from KU Leuven, Leuven, Belgium. under the supervision of Professors Herman Ramon and Josse De Baerdemaeker.

Since 2010 he is an Assistant Professor at the Biosystems Department of KU Leuven, where he leads a group focusing on technology for the AgroFood chain. His main research interests include agricultural automation and robotics, chemometrics, light transport modelling and optical characterisation of biological materials. He has supervised more than 10 PhDs and is (co-)author of over 110 peer reviewed journal articles (ISI)."
\end{IEEEbiography}

\clearpage
\end{document}